\documentclass[letterpaper]{article} 
\usepackage{aaai2027}  
\usepackage[hyphens]{url}  
\usepackage{graphicx} 
\usepackage{natbib}  
\usepackage{caption} 
\usepackage{algorithm}
\usepackage{algorithmic}
\definecolor{mygreen}{RGB}{0,176,80}

\definecolor{grouprow}{gray}{0.88}
\definecolor{subrow}{gray}{0.93}
\usepackage{xcolor}
\definecolor{myyellow}{HTML}{F5E39C}
\usepackage[table]{xcolor}
\usepackage{booktabs}
\usepackage{multirow}
\usepackage{array}
\usepackage{amsmath}
\usepackage{amssymb}
\usepackage{newfloat}
\usepackage{listings}
\DeclareCaptionStyle{ruled}{labelfont=normalfont,labelsep=colon,strut=off} 
\floatstyle{ruled}
\newfloat{listing}{tb}{lst}{}
\floatname{listing}{Listing}

\usepackage{booktabs}

\nocopyright

\title{LAST: The Last Query Token Guides Visual Token Pruning for Edge–Cloud Collaborative MLLM Inference}

\title{LAST: The Last Query Token Guides Visual Token Pruning for Edge–Cloud Collaborative MLLM Inference}
\author {
    Feng Yang\textsuperscript{\rm 1}\equalcontrib,
    Xinrui Ju\textsuperscript{\rm 1}\equalcontrib,
    Keyang Zhang\textsuperscript{\rm 1},
    Xiandong Meng\textsuperscript{\rm 2},
    Rongqun Lin\textsuperscript{\rm 1},\\
    Howard Leung\textsuperscript{\rm 1},
    Shiqi Wang\textsuperscript{\rm 1},
    Haoliang Li\textsuperscript{\rm 1},
    Chris Xing Tian\textsuperscript{\rm 2}\corresponding
}
\affiliations {
  \textsuperscript{\rm 1}City University of Hong Kong\\
    \textsuperscript{\rm 2}Peng Cheng Laboratory\\
    \{fenyang6-c, xinruiju2-c\}@my.cityu.edu.hk,  \{shiqwang, haoliali, howard\}@cityu.edu.hk, \\mengxd@pcl.ac.cn, txsing@live.com
}

\begin{document}

\maketitle

\begin{abstract}
\begin{quote}
Multimodal foundation models are reshaping edge--cloud visual intelligence from task-specific feature pipelines into token-based interfaces, where edge devices encode visual inputs into tokens for a general-purpose cloud MLLM. However, dense visual-token sequences
increase cloud-side inference costs. Existing pruning methods mainly target centralized inference: vision-driven methods can operate before cloud execution but are typically query-agnostic, whereas query-guided methods often rely on internal states of the target MLLM and cannot determine token relevance before transmission. Compact guidance models offer an alternative, but existing designs may require costly attention aggregation or auxiliary generation.
We propose LAST, a training-free framework for query-dependent visual token pruning in edge--cloud collaborative MLLM inference. LAST uses a compact edge-side VLM as a guidance proxy and derives a lightweight importance signal from the last query token’s attention to visual tokens. Under causal attention, the last query token can attend to the full visual sequence and the entire query context, enabling query-aware pruning without cloud-model access, autoregressive generation, or costly aggregation over multiple query positions. LAST then retains a diverse set of query-relevant visual tokens under a fixed token budget.
We evaluate LAST on 11 multimodal benchmarks under multiple token budgets against pruning methods with different guidance strategies. Experiments show that LAST consistently achieves the strongest performance, preserving 95.4\% of the full-token accuracy while retaining only 12.5\% of the visual tokens, with low edge-side selection overhead and reduced cloud-side computation.
\end{quote}
\end{abstract}
\section{Introduction}
\label{sec:intro}
Edge-cloud visual intelligence has long relied on a division of labor
between edge perception and cloud analysis. In representative systems
such as the digital retina, edge devices extract compact visual features
and transmit them to cloud servers, where task-specific models perform
higher-level analysis~\citep{gao2021digitalretina,lou2019digitalretina}.
Although this design avoids transmitting raw visual data, its intermediate
representations and cloud models are typically specialized for predefined
tasks.

\begin{figure}[t]
    \centering
    \includegraphics[width=1\linewidth]{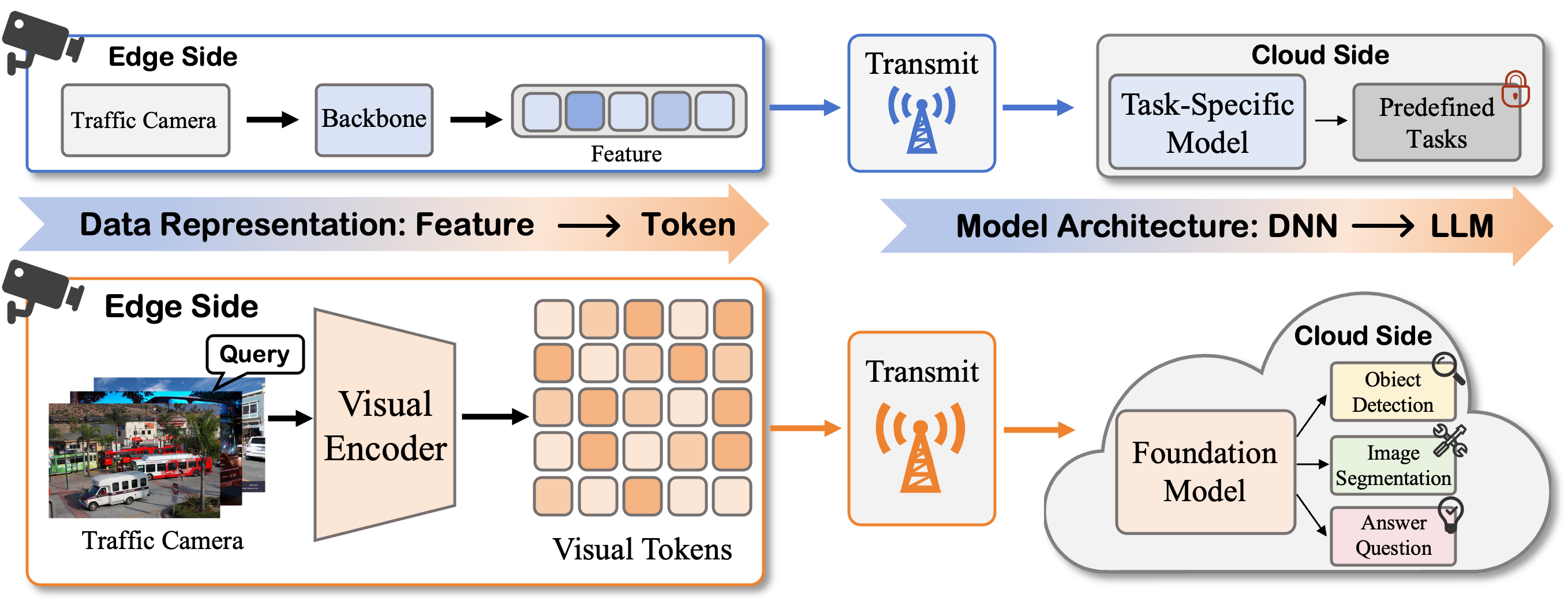}
    \caption{Two paradigms of edge-cloud visual intelligence: task-oriented cloud inference based on edge features, and unified multimodal inference based on visual tokens.
    }
    \label{tab:vis}
\end{figure}

The emergence of multimodal foundation
models~\citep{achiam2023gpt4,li2023mllm,bai2023qwenvl} motivates a more
general architecture: edge devices encode visual observations into
tokens and transmit them to a general-purpose cloud MLLM, which processes
these tokens through a shared reasoning backbone to support diverse
downstream tasks. As illustrated in Figure~\ref{tab:vis}, this evolution
shifts the edge--cloud interface from task-specific features toward
reusable visual tokens, connecting distributed perception with
general-purpose multimodal reasoning.

This token-based interface, however, introduces a new efficiency bottleneck. Vision encoders usually preserve dense patch-level representations, many of which are visually redundant or weakly related to the current query~\citep{liu2023improved,bai2023qwenvl, Visionzip}. Processing the complete visual sequence lengthens the multimodal context and substantially increases cloud-side computation. Visual token pruning is therefore essential for scalable edge-cloud MLLM inference. Importantly, this setting raises a different question from conventional token pruning: rather than deciding which tokens can be removed after the target model has processed the visual input, the edge must determine which visual evidence should be retained before the cloud model observes the full sequence.

Existing visual token pruning methods are largely designed for centralized inference and do not fully meet edge–cloud requirements. Vision-driven methods use visual attention, saliency, or feature similarity~\citep{Visionzip, VisPruner}, but their query-agnostic selection may discard visually inconspicuous yet answer-critical regions. Query-guided methods exploit text–visual interactions in the target VLM~\citep{FastV, SparceVLM}, but obtain relevance signals only after processing the full visual sequence. Methods incorporating diversity or coverage~\citep{CoIn, PruneSID} also assume co-located selection and inference. Thus, direct deployment in edge–cloud settings either delays guidance until cloud execution or requires placing the target MLLM on the edge.

A compact edge-side model offers a possible way to obtain query-conditioned guidance without accessing the cloud MLLM. Prior small-to-large approaches have demonstrated that a smaller VLM can guide token pruning for a larger model~\citep{SGP}. However, existing designs may require aggregation over multiple textual or generated positions, autoregressive generation, or repeated visual encoding. Such operations introduce additional edge-side computation and weaken the benefit of collaborative inference. This raises a key question: \emph{Can query-conditioned visual relevance be estimated at the edge using a lightweight signal that captures the complete image-query context?}

We answer this question with LAST, a training-free framework for query-aware visual token pruning in edge-cloud collaborative MLLM inference. Using a compact edge-side VLM as a guidance proxy, LAST requires only a decoding-free prefill pass. Our key observation is that, under causal attention, the last query token attends to the complete visual prefix and the entire preceding query context, so its attention to the visual tokens serves as a concise query-conditioned relevance signal—without aggregating all query positions, generating a preliminary answer, or accessing the cloud model's internal states. Under a fixed budget, LAST uses this signal to retain a diverse set of query-relevant visual tokens, avoiding redundant allocation to neighboring or semantically similar regions.

We instantiate LAST on scale-varied InternVL models~\citep{InterVL2}, whose shared vision encoder provides a compatible token interface: selected visual tokens are consumed directly by the cloud model through its native connector, without additional alignment or retraining. LAST thus performs query-conditioned selection entirely at the edge while reducing the visual sequence length and cloud-side computation.

We evaluate LAST on diverse multimodal benchmarks under multiple token budgets against query-agnostic, query-dependent, and compact-model pruning baselines. It consistently achieves favorable accuracy-efficiency trade-offs with low edge-side overhead and reduced cloud-side computation. Additional experiments on the LLaVA family~\citep{llava} confirm that the proposed edge-cloud pruning paradigm generalizes beyond InternVL, and system-level measurements further verify its minimal selection overhead.

Our main contributions are summarized as follows:
\begin{itemize}
    \item We formulate visual token pruning for edge-cloud collaborative MLLM inference as a pre-cloud selection problem, in which the pruning mechanism must be query-aware, independent of cloud-model internal states, and sufficiently efficient for edge execution.

    \item We propose LAST, a training-free framework that uses attention from the last query token of a compact edge-side VLM as a concise query-conditioned signal. LAST requires neither autoregressive answer generation nor aggregation over the complete query sequence.

    \item We combine last-query-token relevance with feature-space diversity and conduct extensive evaluations across 11 multimodal benchmarks, multiple token budgets, different pruning-guidance strategies, and two edge-cloud model families, demonstrating favorable accuracy-token and accuracy-computation trade-offs.
\end{itemize}

\section{Related Work}
\label{sec:Related Work}

\subsection{Edge-Cloud Collaborative Vision-Language Inference}

Collaborative edge–cloud vision systems distribute perception and higher-level analysis across resource-constrained devices and powerful cloud servers. Early systems mainly relied on feature-level communication: frameworks such as Digital Retina extract compact visual features at the edge and transmit them to task-specific cloud models, reducing raw-data transmission costs~\citep{gao2021digitalretina,lou2019digitalretina}, but both the transmitted representations and cloud models are typically specialized for predefined tasks. Multimodal foundation models enable a more general interface in which edge-produced visual tokens are consumed by a general-purpose cloud MLLM through a shared multimodal reasoning backbone~\citep{achiam2023gpt4,li2023mllm,bai2023qwenvl}, though this flexibility introduces substantial communication and cloud-side computation costs due to dense visual-token sequences. To address this, some studies investigate how edge–cloud computation should be divided or where requests should be executed: hybrid SLM-LLM systems and CE-CoLLM coordinate models of different capacities, TMO studies adaptive offloading for multimodal, multitask, and multi-turn interactions, and INAR-VL routes requests according to input characteristics and deployment conditions~\citep{HybridSLMLLM,CECoLLM,TMO,INARVL}. These methods operate at the model, request, or execution level, but do not determine which visual evidence within an input should be transmitted to the cloud.

\begin{figure*}[t]
    \centering
    \includegraphics[width=\textwidth]{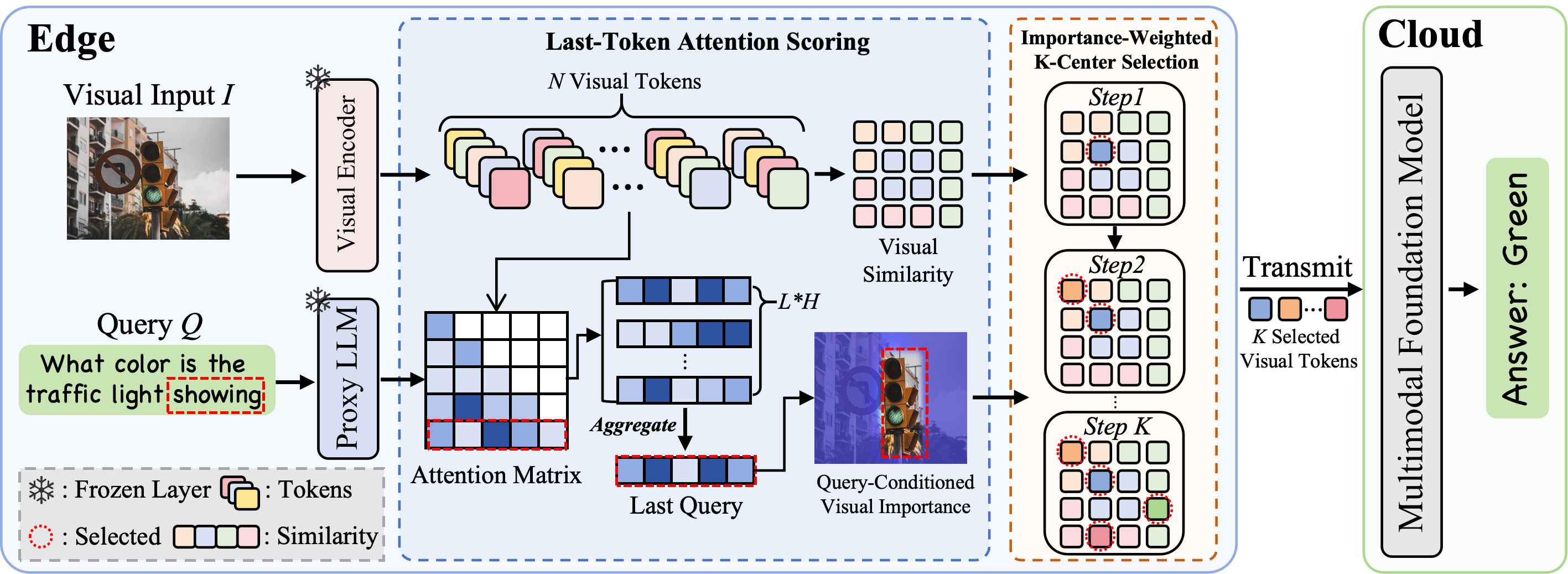}
    \caption{\textbf{Overview of LAST.} A compact edge-side LLM
    performs a single decoding-free prefill pass and uses the last query
    token's attention to estimate query-conditioned visual importance.
    Tokens are then selected by importance-weighted $k$-center selection. The selected features are then transmitted to the cloud model for final response generation.}
    \label{fig:method-overview}
\end{figure*}

Another line of research directly reduces visual communication. Collaborative Edge-to-Server Inference selectively requests detail-preserved regions when an initial cloud prediction is uncertain, while Progressive Semantic Communication compresses visual tokens into progressively refinable representations for transmission under varying network conditions~\citep{song2025collaborative,hsu2026progressive}. These methods optimize the amount or fidelity of transmitted visual information, but do not formulate the problem as query-conditioned selection of the native visual-token sequence entirely before cloud inference. In contrast, LAST performs pre-transmission token selection at the edge: a compact VLM estimates the relevance of candidate visual tokens to the current query, and only a fixed-budget subset is transmitted for final inference by the cloud MLLM. Thus, LAST complements computation placement and representation-compression approaches by reducing the visual-token payload at the edge-cloud interface without accessing cloud-model internal states or introducing a learned compression module.

\subsection{Visual Token Pruning for Multimodal LLMs}

Vision-driven pruning methods estimate visual-token importance from visual salience and redundancy. VisionZip retains dominant tokens and merges contextual information, while VisPruner combines visual-encoder attention with similarity-based redundancy removal~\citep{Visionzip,VisPruner}. Because these methods are largely query-agnostic, they may discard less salient but question-critical evidence. SCoRe, CoIn, and PruneSID improve subset representativeness through feature coverage or diversity~\citep{SCoRe,CoIn,PruneSID}, but still do not ensure the preservation of query-relevant and complementary evidence before transmission.

Language-guided methods derive visual-token importance from text–visual interactions in the target VLM. FastV uses early-layer attention to prune tokens in later layers, while SparseVLM progressively sparsifies the visual sequence using relevant text tokens and self-attention matrices~\citep{FastV,SparceVLM}. QuietPrune enables earlier reduction through a learned query-to-vision adapter~\citep{QuietPrune}. However, FastV and SparseVLM require the full visual sequence to enter the cloud model, whereas QuietPrune relies on target-specific training. Thus, they do not support training-free visual-token selection before cloud-side MLLM execution.

Compact-model guidance enables query-aware pruning before visual tokens reach the cloud MLLM. IF-Prune equips a small VLM with a learned information-bottleneck module to filter query-irrelevant tokens~\citep{IFPrune}. SGP instead aggregates attention from query tokens and autoregressively generated tokens in a compact VLM to construct a patch-wise importance map aligned with the image grid; the target model must still encode the image with its own vision encoder and apply the transferred map for pruning~\citep{SGP}. LAST instead derives importance from only the last query token in a single decoding-free prefill pass and directly transmits the selected edge-encoded tokens, yielding a lightweight pre-transmission pipeline without all-query-token aggregation, autoregressive generation, or repeated visual encoding.

\section{Method}
\label{sec:Method}

\subsection{Overview}
LAST performs query-aware visual-token selection entirely at the edge
before transmitting visual tokens to the cloud. As illustrated in
Figure~\ref{fig:method-overview}, the image is encoded once by a shared
vision encoder, while a compact edge-side VLM performs a single prefill
pass over the complete multimodal prompt without autoregressive
generation. From this pass, LAST extracts a compact query-conditioned
attention signal to estimate visual-token importance and retain a
non-redundant subset under a fixed token budget. Only the selected
visual tokens are transmitted to the cloud, where the cloud model generates the final response.
All model parameters remain frozen, making the complete procedure
training-free.

\subsection{Edge--Cloud Token Interface}
\label{sec:method-design}

We consider an edge--cloud inference system comprising a compact
guidance model $\mathcal{M}_e$ at the edge and a more capable target model $\mathcal{M}_c$ in the cloud. Given an image $I$, the shared vision encoder $E$ produces $N$ visual
tokens
\begin{equation}
\mathbf{X}=E(I)
=[\mathbf{x}_1,\ldots,\mathbf{x}_N]
\in\mathbb{R}^{N\times d_v},
\end{equation}
where $E$ includes the shared vision backbone and spatial tokenization
operations preceding the model-specific connectors. The edge and cloud
models use exactly the same encoder parameters and visual-token layout,
but separate token-wise connectors, $\phi_e$ and $\phi_c$, that preserve
token count and order while projecting the shared encoder tokens into
their respective language-model embedding spaces. This one-to-one
interface allows tokens selected at the edge to be consumed directly by
the cloud connector without feature alignment or image re-encoding.

For a retention ratio $\rho\in(0,1]$, LAST selects an index set
\begin{equation}
\mathcal{S}\subseteq[N],
\qquad
|\mathcal{S}|=K=\lceil\rho N\rceil,
\end{equation} before transmission. The edge model uses $\mathbf{H}_e=\phi_e(\mathbf{X})$ only to estimate the relevance of the encoder visual tokens to the current query. The selected encoder tokens $\mathbf{X}_{\mathcal{S}}$ are transmitted and subsequently projected by the cloud connector $\phi_c$. Reducing the sequence from $N$ to $K$ tokens shortens the visual sequence processed by the cloud connector and language model by a factor of $\rho$.

\subsection{Last-Query-Token Attention Scoring}
\label{sec:method-lasttoken}

LAST extracts query-conditioned visual importance from the last query position of the assembled multimodal sequence, which is conditioned on the complete preceding multimodal context. Unlike approaches that aggregate multiple textual states or require autoregressive decoding, LAST uses attention from a single query position across all layers and heads to obtain a compact relevance signal in one prefill pass.

Given an image and query pair $(I,Q)$, the edge-side visual encoder first produces visual tokens:
\begin{equation}
\mathbf{X}=[\mathbf{x}_1,\ldots,\mathbf{x}_N],
\end{equation}
which are projected into the multimodal embedding space:
\begin{equation}
\mathbf{H}_e=\phi_e(\mathbf{X}).
\end{equation}
After inserting the visual embeddings into the textual query following the standard multimodal input construction process, we obtain the edge-model input:
\begin{equation}
\mathbf{Z}_e
=\mathcal{P}_e(\mathbf{H}_e,Q)
=[\mathbf{z}_1,\ldots,\mathbf{z}_P],
\end{equation}
where $\mathcal{P}_e$ denotes the multimodal input assembly operation. The visual tokens occupy positions $\{p_1,\ldots,p_N\}$, and $p_\star$ denotes the position of the last query token.

At transformer layer $\ell\in\{1,\ldots,L\}$ and attention head $h\in\{1,\ldots,H\}$, let
\begin{equation}
\mathbf{A}^{(\ell,h)}
\in\mathbb{R}^{P\times P}
\end{equation}
denote the softmax-normalized causal self-attention matrix. Due to causal masking, the last query token can attend to all preceding visual tokens and textual context, allowing its attention distribution over visual tokens to capture query-dependent relevance. We therefore aggregate the attention weights from this single query position across all layers and heads:
\begin{equation}
s_i
=
\sum_{\ell=1}^{L}
\sum_{h=1}^{H}
\mathbf{A}^{(\ell,h)}[p_\star,p_i],
\qquad i\in[N].
\end{equation}
The aggregated scores are normalized over the visual sequence:
\begin{equation}
\widetilde{w}_i
=
\frac{s_i}{\sum_{j=1}^{N}s_j},
\qquad
\sum_{i=1}^{N}\widetilde{w}_i=1.
\end{equation}
The resulting relevance distribution
$\widetilde{\mathbf{w}}\in\mathbb{R}_{\geq0}^{N}$
serves as the query-conditioned importance signal for subsequent token selection.

Compared with methods that aggregate multiple textual attention rows, LAST reduces the score-aggregation overhead. Given the required attention weights, aggregating scores from one query position requires $\mathcal{O}(LHN)$ operations, compared with $\mathcal{O}(LHMN)$ for $M$ textual positions. This comparison concerns only score aggregation; the costs of transformer prefill and attention extraction are included in the measured edge-stage latency. In practice, LAST reads the required attention entries from the attention tensors returned during the same prefill pass, without invoking an additional autoregressive generation stage.

\begin{table*}[ht]
\centering
\renewcommand{\arraystretch}{1.15}
\setlength{\tabcolsep}{3pt}
\resizebox{\textwidth}{!}{
\begin{tabular}{l ccccccccccc | cc}
\toprule
\textbf{Method} & \textbf{ChartQA} & \textbf{DocVQA} & \textbf{VQA\textsuperscript{Text}} & \textbf{VQA\textsuperscript{V2}} & \textbf{GQA} & \textbf{VizWiz} & \textbf{SQA\textsuperscript{IMG}} & \textbf{POPE} & \textbf{MME} & \textbf{MMB\textsuperscript{CN}} & \textbf{MM-Vet} & \textbf{Avg.} & \textbf{Rel.} \\
\midrule
Full Tokens & 73.0 & 89.7 & 77.5 & 80.1 & 63.6 & 57.6 & 94.8 & 87.8 & 2222 & 84.9 & 32.3 & 74.6 & 100.0\% \\

\rowcolor{grouprow}
\multicolumn{14}{c}{\textit{Retain 62.5\% Tokens} ($\boldsymbol{\downarrow}$ \textbf{\textit{37.5\%}})} \\
Random & 61.6 & 66.7 & 68.5 & 79.5 & 63.1 & 57.1 & 93.3 & 87.2 & 2175 & 84.2 & 29.0 & 69.8 & 93.6\% \\
\rowcolor{subrow}
\multicolumn{14}{l}{\textit{Query-Agnostic}} \\
VisionZip \textsc{(CVPR25)}   & 65.9 & 75.3 & 72.5 & 79.3 & 62.8 & 55.5 & 93.0 & 88.0 & 2186 & 83.7 & 29.6 & 71.3 & 95.3\% \\
VisPruner \textsc{(ICCV25)}   & 64.1 & 75.0 & 69.9 & 79.4 & 63.1 & 55.6 & 93.5 & 88.3 & 2207 & 83.9 & 26.2 & 70.7 & 94.1\% \\
\rowcolor{subrow}
\multicolumn{14}{l}{\textit{Query-Dependent}} \\
FastV$^\dagger$ \textsc{(ECCV24)}       & 69.8 & 84.4 & 77.2 & 79.8 & 63.5 & 57.6 & 94.4 & 88.1 & 2192 & 84.7 & 30.1 & 73.5 & 98.2\% \\
SparseVLM$^\dagger$ \textsc{(ICML25)}   & 69.4 & 84.4 & 77.0 & 79.9 & 63.6 & 57.6 & 94.6 & 88.1 & 2179 & 84.7 & 29.6 & 73.3 & 98.0\% \\
SGP \textsc{(CVPR25)}         & 68.6 & 84.6 & 77.0 & 79.4 & 63.5 & 57.5 & 94.6 & 88.1 & 2190 & 84.8 & 30.2 & 73.4 & 98.0\% \\
\textbf{LAST (Ours)}          & 70.9 & 89.6 & 77.4 & 79.8 & 63.5 & 57.0 & 94.5 & 88.0 & 2215 & 84.9 & 31.6 & \textbf{74.2} & \textbf{99.3\%} \\

\rowcolor{grouprow}
\multicolumn{14}{c}{\textit{Retain 37.5\% Tokens} ($\boldsymbol{\downarrow}$ \textbf{\textit{62.5\%}})} \\
Random & 46.2 & 44.5 & 55.7 & 77.4 & 62.5 & 55.9 & 90.2 & 86.2 & 2115 & 83.1 & 23.1 & 63.7 & 85.0\% \\
\rowcolor{subrow}
\multicolumn{14}{l}{\textit{Query-Agnostic}} \\
VisionZip \textsc{(CVPR25)}   & 57.9 & 60.0 & 64.2 & 77.4 & 61.6 & 54.4 & 91.1 & 87.7 & 2140 & 83.0 & 24.3 & 67.1 & 89.3\% \\
VisPruner \textsc{(ICCV25)}   & 59.1 & 57.0 & 59.8 & 77.1 & 61.8 & 54.0 & 91.4 & 87.4 & 2157 & 83.5 & 22.3 & 66.4 & 88.1\% \\
\rowcolor{subrow}
\multicolumn{14}{l}{\textit{Query-Dependent}} \\
FastV$^\dagger$ \textsc{(ECCV24)}       & 64.6 & 83.4 & 76.2 & 79.5 & 63.4 & 56.9 & 94.6 & 88.1 & 2203 & 84.6 & 29.3 & 72.7 & 97.0\% \\
SparseVLM$^\dagger$ \textsc{(ICML25)}   & 64.7 & 82.9 & 76.3 & 79.5 & 63.4 & 56.3 & 94.5 & 88.1 & 2199 & 84.6 & 29.3 & 72.5 & 96.9\% \\
SGP \textsc{(CVPR25)}         & 66.6 & 84.1 & 76.5 & 79.5 & 63.3 & 56.7 & 94.5 & 88.2 & 2200 & 84.5 & 30.0 & 73.0 & 97.5\% \\
\textbf{LAST (Ours)}          & 69.4 & 88.4 & 77.3 & 79.5 & 63.5 & 56.5 & 93.7 & 88.4 & 2212 & 84.1 & 32.4 & \textbf{73.8} & \textbf{99.0\%} \\

\rowcolor{grouprow}
\multicolumn{14}{c}{\textit{Retain 12.5\% Tokens} ($\boldsymbol{\downarrow}$ \textbf{\textit{87.5\%}})} \\
Random & 26.0 & 21.6 & 33.7 & 71.0 & 58.5 & 53.3 & 85.4 & 82.5 & 1960 & 79.5 & 16.2 & 54.4 & 72.0\% \\
\rowcolor{subrow}
\multicolumn{14}{l}{\textit{Query-Agnostic}} \\
VisionZip \textsc{(CVPR25)}   & 41.0 & 33.0 & 45.2 & 70.9 & 56.9 & 50.0 & 88.8 & 84.7 & 1883 & 79.5 & 19.4 & 57.8 & 76.8\% \\
VisPruner \textsc{(ICCV25)}   & 37.3 & 27.1 & 39.1 & 68.5 & 54.9 & 47.8 & 87.2 & 82.2 & 1902 & 75.9 & 18.5 & 55.2 & 73.1\% \\
\rowcolor{subrow}
\multicolumn{14}{l}{\textit{Query-Dependent}} \\
FastV$^\dagger$ \textsc{(ECCV24)}       & 45.4 & 74.3 & 71.9 & 78.2 & 61.7 & 53.4 & 90.4 & 88.0 & 2119 & 84.3 & 24.7 & 68.0 & 90.2\% \\
SparseVLM$^\dagger$ \textsc{(ICML25)}   & 47.8 & 73.4 & 72.8 & 78.4 & 62.1 & 53.6 & 91.1 & 88.1 & 2160 & 83.8 & 23.3 & 68.3 & 90.4\% \\
SGP \textsc{(CVPR25)}         & 56.2 & 80.4 & 74.4 & 78.2 & 61.7 & 53.9 & 90.3 & 88.1 & 2099 & 84.3 & 26.9 & 69.9 & 93.0\% \\
\textbf{LAST (Ours)}          & 64.8 & 83.2 & 76.2 & 78.4 & 62.7 & 54.6 & 89.7 & 88.2 & 2097 & 83.2 & 29.5 & \textbf{71.4} & \textbf{95.4\%} \\

\bottomrule
\end{tabular}
}

\caption{\textbf{Accuracy on 11 benchmarks under the InternVL2-1B/8B edge--cloud setting.}
Full Tokens is the unpruned baseline. Avg. is the macro-average (MME normalized to 2800), and Rel. is the relative performance w.r.t. Full Tokens. Best Avg./Rel. are \textbf{bold}. $^\dagger$ denotes proxy-LLM adaptation for pre-transmission pruning.}
\label{tab:main}
\end{table*}

\subsection{Pre-Transmission Token Selection}
\label{sec:method-selection}

Directly retaining the $K$ largest importance scores may allocate much of the limited token budget to visually similar features. To balance query relevance and visual diversity, we adopt an importance-weighted greedy $k$-center selection rule, where the query-conditioned importance scores $\widetilde{\mathbf{w}}$ guide a greedy coverage-based selection.

We first normalize each encoder feature as
$\bar{\mathbf{x}}_i=\mathbf{x}_i/\|\mathbf{x}_i\|_2$ and initialize the
selected set with the most query-relevant token:
\begin{equation}
j_1=\arg\max_{i\in[N]}\widetilde{w}_i,
\qquad
\mathcal{S}_1=\{j_1\}.
\end{equation}
At iteration $t\geq2$, the current coverage distance of an unselected
token is computed as
\begin{equation}
\delta_i^{(t-1)}
=
\min_{j\in\mathcal{S}_{t-1}}
(1-\bar{\mathbf{x}}_i^\top\bar{\mathbf{x}}_j).
\end{equation}
We then select the token maximizing the product of query relevance and
coverage gain:
\begin{equation}
j_t
=
\arg\max_{i\notin\mathcal{S}_{t-1}}
\widetilde{w}_i\,\delta_i^{(t-1)},
\qquad
\mathcal{S}_t
=
\mathcal{S}_{t-1}\cup\{j_t\}.
\end{equation}
This multiplicative objective suppresses tokens that are either
irrelevant to the query or redundant with already selected tokens,
requiring both high relevance and large marginal coverage gain.

For dynamically tiled inputs, visual tokens from all local tiles and the
thumbnail are flattened into a unified candidate pool and compete under
the same global budget, without per-tile quotas or mandatory token
retention. Text tokens are never pruned. After $K$ iterations, the
selected indices are restored to their original order:
\begin{equation}
\mathbf{X}_{\mathcal{S}}
=
[\mathbf{x}_i]_{i\in\operatorname{sort}(\mathcal{S}_K)}.
\end{equation}
Only $\mathbf{X}_{\mathcal{S}}$ is transmitted to the cloud. The cloud
model directly applies its native connector
$\mathbf{H}_{\mathcal{S}}=\phi_c(\mathbf{X}_{\mathcal{S}})$ and generates
the final response, without modifying the cloud-side architecture.

With incremental updates of $\delta_i^{(t)}$, the selection process requires $\mathcal{O}(KNd_v)$ distance computations and avoids constructing an $N\times N$ pairwise-distance matrix.

\section{Experiment}
\label{sec:Experiment}

\begin{table}[t]
\centering
\renewcommand{\arraystretch}{0.95}
\setlength{\tabcolsep}{2.5pt}
\resizebox{\columnwidth}{!}{
\begin{tabular}{l cccc}
\toprule
\textbf{Method} & \textbf{Agg. (ms)} & \textbf{Edge (ms)} & \textbf{FLOPs (T)} & \textbf{Acc.} \\
\midrule
Full Tokens & -- & 118.55 & 32.30 & 73.0 \\
\rowcolor{grouprow}
\multicolumn{5}{c}{\scriptsize\textit{Retain 62.5\% Tokens} ($\boldsymbol{\downarrow}$ \textbf{\textit{37.5\%}})} \\
FastV$^\dagger$     & \underline{3.96}  & \textbf{235.02} & 24.02 & \underline{69.8} \\
SparseVLM$^\dagger$ & 51.06            & 279.57          & 24.02 & 69.4 \\
SGP       & 6.23             & 285.56          & 24.09 & 68.6 \\
\textbf{LAST (Ours)} 
          & \textbf{2.51}    & \underline{237.92} & 24.04 & \textbf{70.9} \\
\rowcolor{grouprow}
\multicolumn{5}{c}{\scriptsize\textit{Retain 12.5\% Tokens} ($\boldsymbol{\downarrow}$ \textbf{\textit{87.5\%}})} \\
FastV$^\dagger$     & \underline{3.97} & \underline{239.95} & 10.98 & 45.4 \\
SparseVLM$^\dagger$ & 50.87            & 278.17             & 10.98 & 47.8 \\
SGP       & 6.18             & 280.04             & 11.10 & \underline{56.2} \\
\textbf{LAST (Ours)} 
          & \textbf{2.51}    & \textbf{236.43}    & 11.01 & \textbf{64.8} \\
\bottomrule
\end{tabular}
}
\caption{
\textbf{Efficiency comparison of query-dependent pruning methods on ChartQA.}
Agg. denotes token-importance aggregation time, and Edge denotes
edge-stage latency per inference. FLOPs cover the complete pipeline.
Full Tokens denotes unpruned inference without proxy guidance. Best and
second-best results within each budget are shown in \textbf{bold} and
\underline{underlined}.
}
\label{tab:performance}
\end{table}

\begin{figure}[t]
\centering
\includegraphics[width=\linewidth]{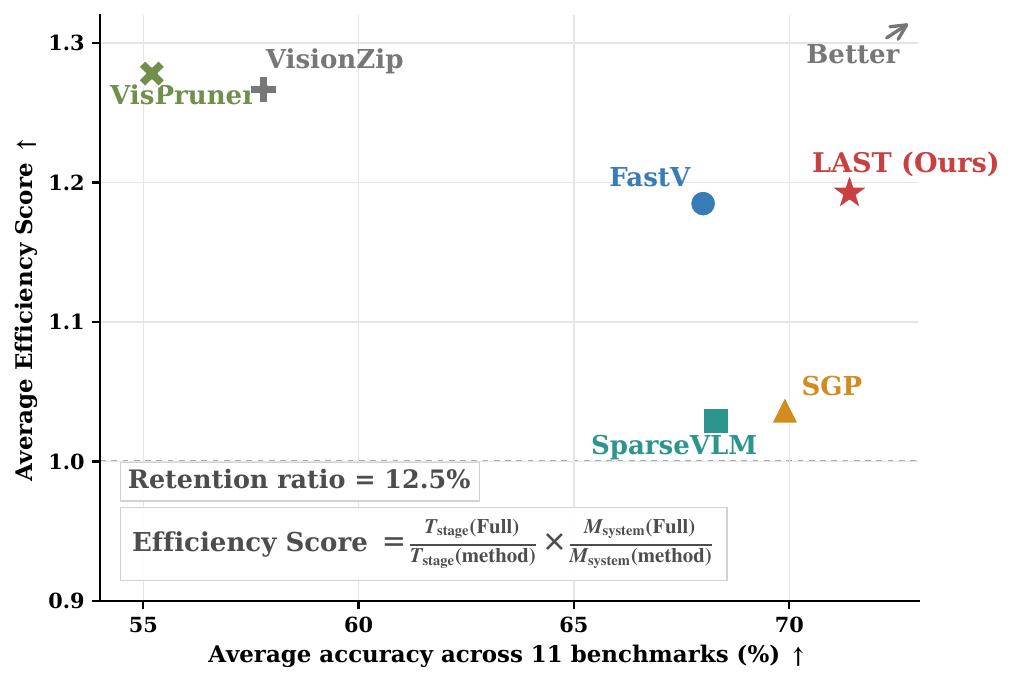}
\caption{\textbf{Accuracy–efficiency trade-off on Edge-Cloud Model.} The average efficiency score jointly captures the prefill-stage speedup and GPU memory reduction relative to the full-token baseline across all datasets.}
\label{fig:Accuracy–efficiency-tradeoff}
\end{figure}

\subsection{Experimental Settings}

\textbf{Models.} We instantiate the default edge-cloud system with an InternViT-300M-448px edge vision encoder, a Qwen2-0.5B-Instruct proxy LLM~\citep{yang2024qwen2}, and an InternLM2.5-7B-Chat cloud LLM, using components from the official InternVL2-1B and InternVL2-8B checkpoints~\citep{InterVL2}. Specifically, the edge vision encoder and proxy LLM, together with their associated connector, are taken from InternVL2-1B, while the cloud LLM and its connector are taken from InternVL2-8B. This decomposition reflects their distinct roles in our framework: the vision encoder produces candidate visual tokens, the lightweight proxy LLM estimates their query-conditioned importance, and the cloud LLM performs final response generation.
The full-token reference uses the same edge-cloud configuration but bypasses the proxy LLM and transmits all visual tokens.
All model components remain frozen, and we follow the official image preprocessing and chat templates. Since InternVL2 uses dynamic resolution, retention ratios are applied to the number of visual tokens in each image. Experiments are conducted on a single NVIDIA A800 80GB GPU, and random selection is averaged over three independent runs.

\noindent \textbf{Datasets.}  We conduct experiments on 11 vision-language benchmarks to comprehensively evaluate our method. These benchmarks span multiple domains: document and chart understanding ChartQA \citep{ChartQA} and DocVQA \citep{Docvqa}, scene text recognition TextVQA \citep{TextVqa}, general visual question answering VQAv2 \citep{VqaV2}, GQA \citep{GQA} and VizWiz \citep{Vizwiz}, science question answering ScienceQA-IMG \citep{ScienceQA}, hallucination evaluation POPE \citep{PoPe}, and multi-modal comprehensive evaluation MME \citep{MME}, MMBench-CN \citep{MMbench} and MM-Vet \citep{Mm-vet}.


\begin{table}[t]
  \centering
  \renewcommand{\arraystretch}{1.12}
  \setlength{\tabcolsep}{2.5pt}
  \resizebox{\columnwidth}{!}{
  \begin{tabular}{l ccccc cc}
  \toprule
  \textbf{Method} & \textbf{VQA\textsuperscript{V2}} & \textbf{GQA} & \textbf{POPE} & \textbf{MME} & \textbf{MM-Vet} & \textbf{Avg.}$\uparrow$ & \textbf{Rel.}$\uparrow$ \\
  \midrule
  Full Tokens  & 78.3 & 63.4 & 85.9 & 1808 & 25.2 & 63.5 & 100.0\% \\

  \rowcolor{grouprow}
  \multicolumn{8}{c}{\textit{Retain 128 / 576 Visual Tokens} ($\boldsymbol{\downarrow}$ \textbf{\textit{77.8\%}})} \\
  VisionZip  & 74.7 & 58.6 & 82.9 & 1769 & 24.2 & 60.7 & 95.7\% \\
  VisPruner  & 74.9 & 58.0 & 84.2 & 1762 & 24.5 & 60.9 & 95.9\% \\
  \textbf{LAST (Ours)} & \textbf{76.8} & \textbf{62.1} & \textbf{86.9} & \textbf{1783} & \textbf{25.2} & \textbf{62.9} & \textbf{99.2\%} \\

  \rowcolor{grouprow}
  \multicolumn{8}{c}{\textit{Retain 64 / 576 Visual Tokens} ($\boldsymbol{\downarrow}$ \textbf{\textit{88.9\%}})} \\
  VisionZip  & 71.6 & 56.8 & 75.8 & 1648 & \textbf{23.3} & 57.3 & 90.2\% \\
  VisPruner  & 71.7 & 56.2 & 79.7 & 1719 & 22.2 & 58.2 & 91.7\% \\
  \textbf{LAST (Ours)} & \textbf{75.1} & \textbf{60.0} & \textbf{87.9} & \textbf{1748} & 22.2 & \textbf{61.5} & \textbf{97.0\%} \\

  \rowcolor{grouprow}
  \multicolumn{8}{c}{\textit{Retain 32 / 576 Visual Tokens} ($\boldsymbol{\downarrow}$ \textbf{\textit{94.4\%}})} \\
  VisionZip  & 66.0 & 53.8 & 67.2 & 1514 & 20.1 & 52.2 & 82.3\% \\
  VisPruner  & 67.8 & 52.6 & 71.5 & 1537 & \textbf{22.4} & 53.8 & 84.8\% \\
  \textbf{LAST (Ours)} & \textbf{71.0} & \textbf{54.6} & \textbf{84.6} & \textbf{1593} & 19.3 & \textbf{57.3} & \textbf{90.2\%} \\

  \bottomrule
  \end{tabular}
  }
  \caption{\textbf{Cross-model evaluation under the LLaVA-1.5-7B/13B configuration.} Full Tokens denotes the unpruned reference. Avg. is the macro-average across five benchmarks. Rel. is the macro-average of per-benchmark performance relative to Full Tokens. The best results within each token budget are shown in \textbf{bold}.}
  \label{tab:llava_transfer}
\end{table}

\noindent \textbf{Comparison Methods.}
We compare LAST with random selection, two query-agnostic methods, VisionZip~\citep{Visionzip} and VisPruner~\citep{VisPruner}, and three query-dependent methods, FastV~\citep{FastV}, SparseVLM~\citep{SparceVLM}, and SGP~\citep{SGP}. FastV and SparseVLM were originally designed for single-model inference, where visual tokens are pruned within the LLM decoder. To adapt them to pre-transmission pruning, we apply their original attention-based importance estimation and retention rules to the edge-side proxy LLM and transmit the retained tokens to the cloud. SGP already follows a small-to-large scheme and therefore naturally fits our edge-cloud setting. All methods use the same edge vision encoder, cloud LLM, visual inputs, prompts, retention budgets, and decoding configuration, while all query-dependent methods share the same proxy LLM.

\subsection{Main Results}

\textbf{Evaluation Accuracy.}
Table~\ref{tab:main} reports the results across 11 benchmarks under
three visual-token retention budgets. LAST consistently achieves the
best Avg. and Rel. at all budgets. When retaining 62.5\%, 37.5\%,
and 12.5\% of the visual tokens, LAST preserves 99.3\%, 99.0\%,
and 95.4\% of the full-token performance, respectively. Compared
with the strongest competing method at each budget, LAST improves
Rel. by 1.1, 1.5, and 2.4 percentage points, showing that its
advantage becomes more pronounced as the token budget decreases.

The results also highlight the importance of query-dependent
selection. At 12.5\% retention, the best query-agnostic method
achieves an Avg. of only 57.8, whereas SGP and LAST reach 69.9 and
71.4, respectively. LAST is particularly effective on benchmarks
that require identifying localized information from visually dense
inputs. Under the same aggressive budget, it outperforms SGP by
8.6 points on ChartQA, 2.8 points on DocVQA, and 2.6 points on
MM-Vet. Although the best method varies across individual benchmarks,
LAST provides the strongest overall accuracy-token trade-off and
degrades more gracefully under aggressive token reduction.

\noindent \textbf{Efficiency Analysis.}
At a fixed retention ratio, all methods transmit the same number of visual tokens, so their efficiency differences mainly arise from edge-side guidance. As shown in Table~\ref{tab:performance}, LAST requires only 2.51 ms to derive token-importance scores, 36.8\% less than FastV and substantially less than SparseVLM. At 12.5\% retention, LAST achieves both the highest accuracy and the lowest edge latency among the query-dependent methods, outperforming SGP by 8.6 points while being 43.61 ms faster. We further visualize the accuracy–efficiency trade-off in Figure~\ref{fig:Accuracy–efficiency-tradeoff}. VisionZip and VisPruner obtain slightly higher efficiency scores as
their query-agnostic pruning avoids any importance aggregation, but their average accuracy drops substantially, far below competing methods. In contrast, LAST achieves the highest accuracy while retaining a competitive efficiency score. Overall, LAST provides a favorable accuracy-efficiency trade-off under aggressive token reduction.

\noindent \textbf{Cross-Model Evaluation.}
To evaluate whether LAST generalizes beyond the InternVL2 family, we further adopt LLaVA-1.5-7B, the smallest available LLaVA-1.5 variant, as the proxy model and LLaVA-1.5-13B as the cloud model~\citep{llava}. As shown in Table~\ref{tab:llava_transfer}, LAST consistently achieves
the best Avg. and Rel. across all token budgets. It retains 99.2\%, 97.0\%, and 90.2\% of the full-token performance using only 128, 64, and 32 of the original 576 visual tokens, respectively. 
These results demonstrate that LAST transfers effectively to a different MLLM family and visual-token interface.

\begin{figure}[t]
    \centering
    \setlength{\belowcaptionskip}{-10pt}
    \includegraphics[width=1\linewidth]{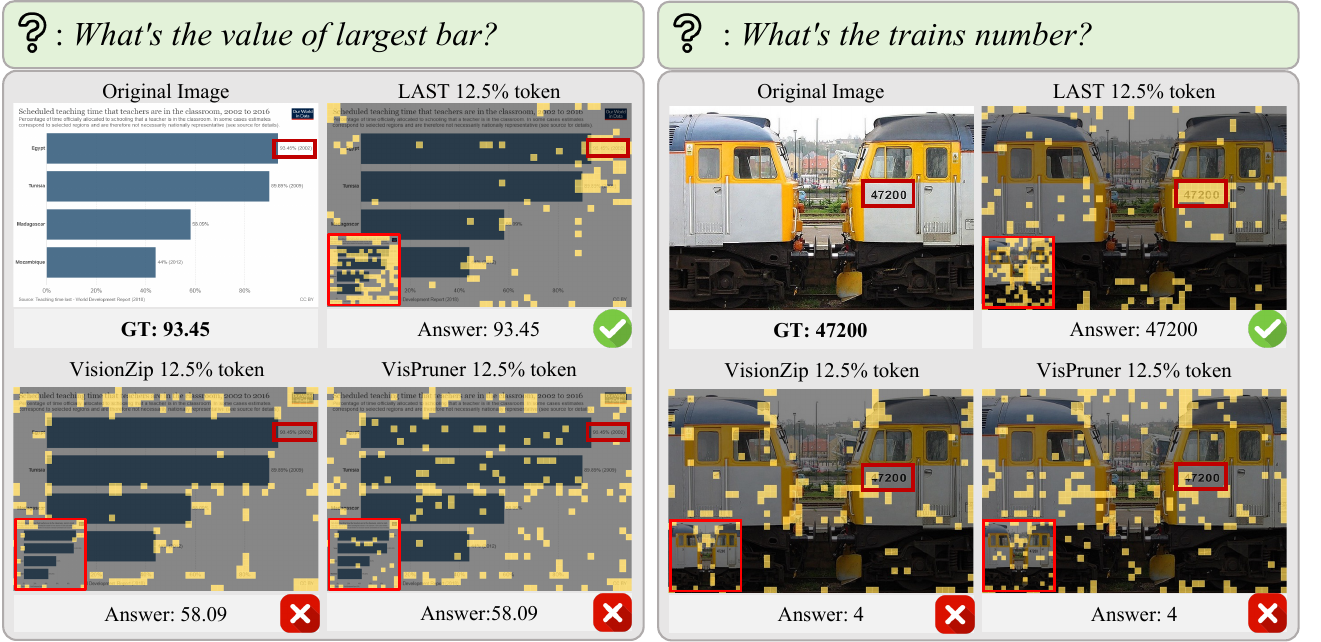}
    \caption{\textbf{Visualization of LAST compared to query-agnostic methods.} 12.5\% retained tokens are highlighted with \textcolor{myyellow}{\rule{1.2ex}{1.2ex}}. Thumbnails employed in InternVL are presented in the left corner.
    }
    \label{fig:qualitative}
\end{figure}

\subsection{Ablation and Analysis}
\noindent \textbf{Qualitative Analysis of Query-Aware Selection.} Figure~\ref{fig:qualitative} visualizes the retained visual tokens on representative ChartQA and DocVQA examples at 12.5\% retention. Under the same limited token budget, VisionZip and VisPruner allocate tokens to regions favored by their query-agnostic criteria but omit compact regions containing the evidence required by the questions. In contrast, LAST concentrates the retained tokens around the relevant chart values and train number, producing the correct answers \(93.45\) and \(47200\), respectively. These examples suggest that query conditioning can help preserve ground-truth evidence when it occupies only a small portion of a visually dense input.


\noindent \textbf{Choosing the Attention Source.}
We compare different attention sources for deriving visual-token
importance in Table~\ref{tab:query_token_ablation}. LAST achieves the highest average performance at all three retention ratios. Averaging all query-token attention rows yields
competitive results, while additionally incorporating generated-answer attention provides further gains in some settings. The former requires aggregation over the full query, whereas the latter additionally requires an autoregressive generation pass. Overall, these results suggest that the last query token provides an effective and compact source of query-conditioned guidance.

\begin{table}[t]
  \centering
  \setlength{\tabcolsep}{3.0pt} 
  \resizebox{\columnwidth}{!}{
  \begin{tabular}{l ccccc}
  \toprule
  \textbf{Attention Source} & \textbf{ChartQA}$\uparrow$ & \textbf{VQAv2}$\uparrow$ & \textbf{MME}$\uparrow$ & \textbf{MM-Vet}$\uparrow$ & \textbf{Avg.}$\uparrow$ \\
  \midrule

  \rowcolor{grouprow}
  \multicolumn{6}{c}{\textit{Retain 62.5\% Tokens}} \\
  First             & 66.7 & 79.3 & 2172 & 31.4 & 63.7 \\
  Middle            & 69.5 & 79.5 & 2186 & 31.5 & 64.6 \\
  All Mean          & 70.2 & 79.6 & \textbf{2218} & 31.1 & 65.0 \\
  All Mean + Answer & 70.1 & 79.7 & 2215 &  31.6    &  65.1    \\
  \textbf{Last (Ours)} & \textbf{70.9} & \textbf{79.8} & 2215 & \textbf{31.6} & \textbf{65.4} \\

  \rowcolor{grouprow}
  \multicolumn{6}{c}{\textit{Retain 37.5\% Tokens}} \\
  First             & 56.1 & 78.6 & 2208 & 26.4 & 60.0 \\
  Middle            & 62.3 & 78.8 & 2193 & 29.8 & 62.3 \\
  All Mean          & 68.2 & 79.3 & 2207 & 27.7 & 63.5 \\
  All Mean + Answer & 69.0 & 79.4 & \textbf{2238} & 31.6   &  65.0    \\
  \textbf{Last (Ours)} & \textbf{69.4} & \textbf{79.5} & 2212 & \textbf{32.4} & \textbf{65.1} \\

  \rowcolor{grouprow}
  \multicolumn{6}{c}{\textit{Retain 12.5\% Tokens}} \\
  First             & 35.5 & 74.8 & 2045 & 19.3 & 50.7 \\
  Middle            & 43.0 & 74.9 & 2041 & 22.6 & 53.3 \\
  All Mean          & 59.6 & 78.3 & 2092 & 26.6 & 59.8 \\
  All Mean + Answer & 64.5 & 78.4 & 2078 & 29.1 & 61.6     \\
  \textbf{Last (Ours)} & \textbf{64.8} & \textbf{78.4} & \textbf{2097} & \textbf{29.5} & \textbf{61.9} \\

  \bottomrule
  \end{tabular}
  }
\caption{\textbf{Ablation on the attention source used to derive
visual-token importance.} For a query containing \(M\) tokens, First, Middle, and Last use the attention rows of \(q_1\), \(q_{\lceil {p_\star}/2\rceil}\), and \(q_{p_\star}\), respectively. All Mean averages the attention rows of all query tokens, while All Mean + Answer additionally includes attention from answer tokens autoregressively generated by the proxy model.}
  \label{tab:query_token_ablation}
\end{table}


\begin{table}[t]
\centering
\renewcommand{\arraystretch}{0.95}
\setlength{\tabcolsep}{6pt}
\resizebox{\columnwidth}{!}{
\begin{tabular}{l c cc}
\toprule
\textbf{Method} &  \textbf{ChartQA (Relaxed Acc.)$\uparrow$} & \textbf{MME (Official Score)$\uparrow$ } \\
\midrule
\rowcolor{grouprow}
\multicolumn{3}{c}{\textit{Retain 12.5\% Tokens}} \\
    Random  & 26.0  & 1960 \\
    LAST (Top-k)  & 60.1 & 2058 \\
    \textbf{LAST (Ours)}    & \textbf{64.8}  & \textbf{2097} \\
\bottomrule
\end{tabular}
}
\caption{\textbf{Selection-strategy ablation at 12.5\% token retention.} Both LAST (Top-$k$) and LAST (Ours) use the last-query-token attention as the token importance score. Top-$k$ directly retains the highest-scoring tokens, while Ours applies the proposed importance-weighted $k$-center selection to balance relevance and diversity.}
\label{tab:performance_comparison}
\end{table}

\noindent \textbf{Effectiveness of the Selection Strategy.}
To isolate the effect of the selection strategy, we compare LAST with random selection and direct Top-$k$ selection using the same last-query-token importance scores. As shown in Table~\ref{tab:performance_comparison}, the proposed importance-weighted k-center selection outperforms Top-$k$ by 4.7 points in accuracy on ChartQA and by 39 in MME score at 12.5\% retention. This suggests that jointly considering relevance and feature diversity uses the limited token budget more effectively than importance-only ranking.


\section{Conclusion}
\label{sec:conclusion}
We propose LAST, a training-free visual-token pruning framework for
collaborative edge--cloud MLLM inference. LAST uses attention from the last
query token of a compact edge-side guidance model to estimate
query-conditioned visual importance before transmission. By combining this
signal with feature-space diversity, LAST preserves task-relevant and
complementary visual evidence while removing redundant tokens under a fixed
budget. Experiments across visual question answering and multimodal
understanding benchmarks demonstrate favorable performance--efficiency
trade-offs under multiple token budgets. The results show that lightweight,
query-aware pre-transmission pruning can reduce visual-token transmission
cloud-side computation while maintaining strong downstream performance,
providing an effective token interface for edge--cloud multimodal inference.


\bigskip

\bibliography{aaai2027}


\end{document}